\documentclass[runningheads]{llncs}
\pdfoutput=1
\usepackage{graphicx}
\usepackage{url}
\usepackage{color}
\usepackage{hyperref}
\usepackage[table,xcdraw]{xcolor}
\usepackage{amsmath}
\usepackage{amssymb}

\usepackage[table,xcdraw]{xcolor}
\usepackage{caption}
\usepackage{graphicx}
\usepackage{booktabs}
\usepackage[normalem]{ulem}
\usepackage{subcaption}
\usepackage{tikz}

\usepackage{framed}
\usepackage{listings}

\usepackage{algorithm}
\usepackage{algpseudocode}
\definecolor{lightblue}{RGB}{240,245,255}
\definecolor{codegray}{rgb}{0.5,0.5,0.5}
\definecolor{ceruleanblue}{rgb}{0.16, 0.32, 0.75}
\definecolor{coolblack}{rgb}{0.0, 0.18, 0.39}
\definecolor{cadmiumorange}{rgb}{0.93, 0.53, 0.18}

\definecolor{darkblue}{rgb}{0,0.08,0.45}

\definecolor{shadecolor}{RGB}{240,245,255}
\usepackage[most]{tcolorbox}

\begin{document}
\title{The GraphNet Zoo: An All-in-One Graph Based Deep Semi-Supervised Framework for Medical Image Classification}
%
%
\author{
Marianne de Vriendt\inst{*1}
Philip Sellars\inst{*2}  \and
Angelica I. Aviles-Rivero\thanks{These three authors contributed equally and hold joint first authorship.}\inst{3}
}
\authorrunning{}
%
\institute{
Nabla Technologies, Paris, France  \email{marianne@nabla.com}    \and
DAMPT and $^3$DPMMS, University of Cambridge, UK  \email{\{ps644,ai323\}@cam.ac.uk}
}
\maketitle              
\begin{abstract}
We consider the problem of classifying a medical image dataset when we have a limited amounts of labels. This is very common yet challenging setting as labelled data is expensive, time consuming to collect and may require expert knowledge. The current classification go-to of deep supervised learning is unable to cope with such a problem setup. However, using semi-supervised learning, one can produce accurate classifications using a significantly reduced amount of labelled data. Therefore, semi-supervised learning is perfectly suited for medical image classification. However, there has almost been no uptake of semi-supervised methods in the medical domain. In this work, we propose an all-in-one framework for deep semi-supervised classification focusing on graph based approaches, which up to our knowledge it is the first time that an approach with minimal labels has been shown to such an unprecedented scale with \textit{medical data}. We introduce the concept of hybrid models by defining a classifier as a combination between an energy-based model and a deep net. Our energy functional is built on the  Dirichlet energy based on the graph p-Laplacian. Our framework includes energies based on the $\ell_1$ and $\ell_2$ norms.  We then connected this energy model to a deep net to generate a much richer feature space to construct a stronger graph. Our framework can be set to be adapted to any complex dataset.
We demonstrate, through extensive numerical comparisons, that our approach readily compete with fully-supervised state-of-the-art techniques for the applications of Malaria Cells, Mammograms  and  Chest X-ray  classification whilst using only 20\% of labels.

\end{abstract}
\section{Introduction}

Deep learning for medical image classification has achieved state-of-the-art results for a variety of medical image classification challenges \cite{baltruschat2019comparison,yao2018weakly,wang2017chestx,shen2019deep} However, state-of-the-art deep learning frameworks rely upon the existence of a representative training set, which often requires  a large number of manually labelled medical images. Collecting labelled data for medical imaging is a time-consuming, expensive and requires domain expertise from trained physicians. Therefore, obtaining such a representative training set is often a barrier to machine learning in the medical domain.

Semi-supervised learning (SSL) techniques have been growing massively in popularity due to the fact they seek to produce an accurate solution whilst using a minimal size label set. SSL techniques seek to use the information present in a large number of unlabelled examples combined with a small number of labelled examples \cite{chapelle2009semi} to obtain a better performance than purely using the labelled samples on their own.  There are several different approaches to SSL which can be split into several board families of methods: \textit{low-density separation} \cite{chapelle_low_density} , \textit{generative models} \cite{gener_exa} and graph based approaches \cite{zhou2004learning,kipf2016semi} . In this paper we will narrow our discussion to graphical techniques due to their flexibility in dealing with different data structures, scalability to large problems and their rigorous mathematical definition.

SSL is perfect for any area which produces large quantities of data but also incurs a large cost associated with labelling. Thus making SSL techniques a perfect candidate for use in the field of medical image classification. However, there has been very limited uptake of semi-supervised learning techniques for use in medical image classification. In this paper, we seek to bridge this technical gap and demonstrate, to our knowledge for the first time, the amazing results that can be obtained by using deep SSL for medical large-scale image classification.

The theoretical foundations of SSL has been studied by the community for years. But it is only recently that deep semi-supervised learning has be a focus of great attention. Several techniques has been proposed including~\cite{tarvainen2017mean,verma2019interpolation}. However, these techniques has been only proven effective for natural images, and the question of how effective they are  on complex datasets such as those coming from the medical domain has not been investigated yet. This is not obvious-- as there are fundamental differences between natural and medical images~\cite{raghu2019transfusion}. To our knowledge, this paper represented the first major exploration of deep semi-supervised learning for large scale medical datasets.

\medskip
\textcolor{darkblue}{\textbf{Our Contributions.}}We propose an all-in-one framework for Deep Semi-Supervised Medical Image Classification framed into a package called GraphNet Zoo.  Our framework works as a hybrid technique that uses an energy model as a core to drive the uncertainty updates through a deep network. Our particular highlights are:

\begin{itemize}
    \item A generalisable framework which is composed of an energy based model and a deep net.  Whilst the embeddings coming from the deep net aim to construct a robust graph, the optimisation model drives the final graph based classifier. Our energy model is based on minimising the Dirichlet energy based on the graph p-Laplacian, which we integrate two cases: $p=2$ and a more robust functional based on the non-local total variation $p=1$. We also show that our approach can plug-and-play any deep net architecture.
    \item We demonstrate, through an extensive experiments and for a  range of complex medical datasets, that our framework can recreate, and in some cases outperform, the performance of supervised methods whilst using only 20\%.
    \item To the best of our knowledge, this is the first time that a deep semi-supervised framework has been applied and been shown to output fantastic performance to several large-scale \textit{medical datasets}.
\end{itemize}

\section{GraphNet Zoo: An All-in-One Framework}
The lack of a large corpus of well-annotated medical data has motivated the developed of new techniques, which need a significantly smaller set of labelled data. Unlike other type of data (e.g. natural images), the complex annotations required in the medical domain advocates for at least a double reading from different experts which is highly subjective and prone to error~\cite{lehman2015diagnostic}. At the algorithmic level, this is reflected in greater label uncertainty that negatively affects the classification task. Therefore, how to rely less in annotated data is of a great interest in the medical domain. The body of literature has explored different alternatives including Transfer Learning e.g~\cite{bar2015chest} and Generative Adversarial Models e.g.~\cite{moradi2015machine} to mitigate somehow the lack of well-annotated medical data. However, the existing algorithmic approaches do not consider the discrepancy between the expert and the ground truth. We address this problem by proposing a graph-based deep semi-supervised framework. We first formalise the problem that we aim to solve.

\begin{figure}[t!]
\centering
\includegraphics[width=1\textwidth]{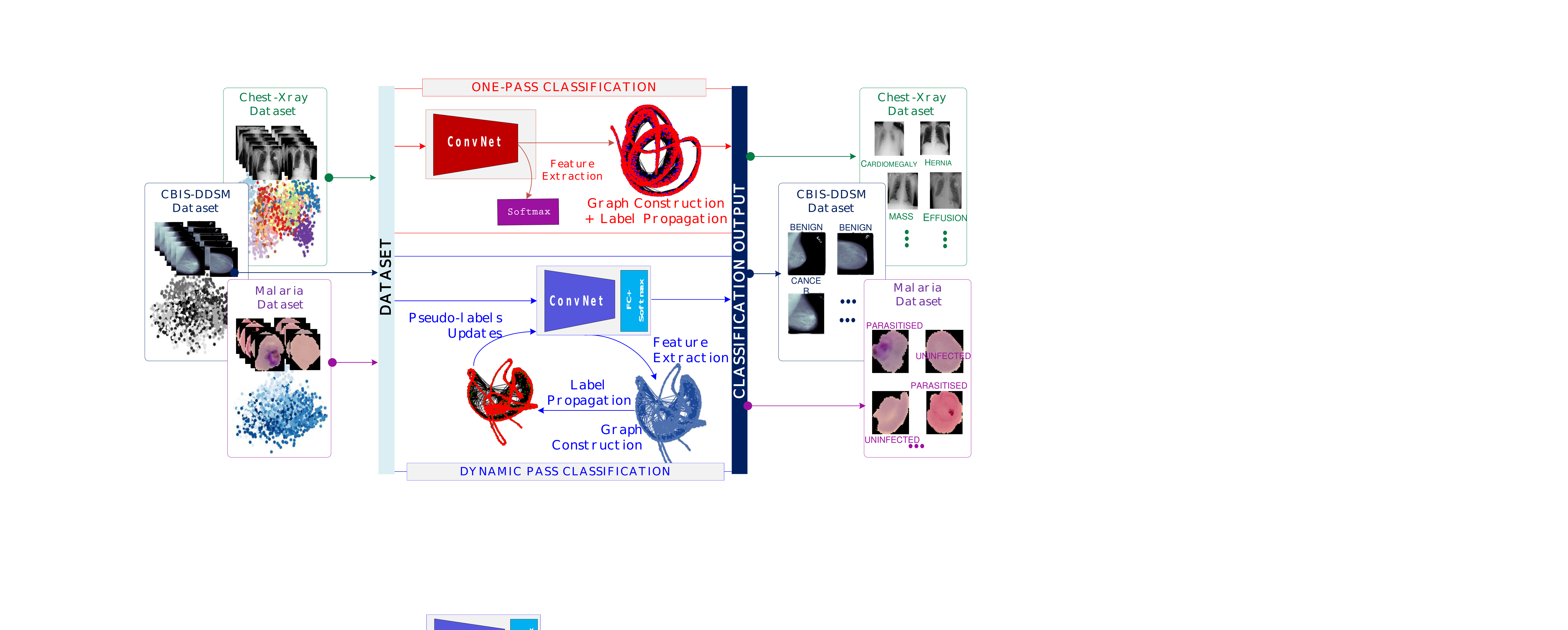}
\caption{Visual description of our proposed GraphZoo framework. We split the different graph based models into two different groups, coloured in red and blue, dependent on whether the algorithm is iterative. No update (red) approaches construct the graph only once and immediately perform graph based operations such as label propagation. Update approaches iteratively construct the graphical representation, using the features extracted from the previous epoch to construct an improved graph.}
\label{fig::teaser}
\end{figure}

\smallskip
\textcolor{darkblue}{\textbf{Problem Statement.}} Assume a set of inputs $X := \{x_1,x_2,...,x_n\}$. For $1 \leq i \leq l$, $x_i$ has a label $y_i \in C := \{1,...,c\}$, where C is a discrete label set for $c$ classes. The labels $y_i$ form a set $Y := \{y_1,y_2,...,y_l\}$. As such we split $X = X_L \cup X_U$ where $X_L := \{x_1,..,x_l\}$ and $X_U := \{x_{l+1},..,x_n\}$. We then seek to  use $X_L$, $X_U$ and $Y_L$ to find an optimal mapping $f : \mathcal{X} \rightarrow  \mathbb{R}^C$, with minimal error, that  can accurately predict the labels $Y_U = \{ y_{l+1},..,y_{u+l} \}$ for the unlabelled points $X_U$ and potential infinitely unseen instances. The mapping $f$ is parameterised by $\theta$ and can be decomposed as $f = \phi \circ \psi$, where $\psi : \mathcal{X}\rightarrow \mathbb{R}^P$ is a feature extractor that maps the input to some feature space of dimension $P$ and $\phi : \mathbb{R}^P \rightarrow \{0,1\}^c$ is the classification function. In the context of this paper $\phi$ will be a graph based classifier.

\smallskip
\textcolor{darkblue}{\textbf{How We Represent the Data?}} For the majority of existing approaches in medical classification, the go-to representation of the the data is thee a standard grid form. In this work, we give a different representation - graphical. Formally, a given dataset can be represented as an undirected weighted graph $\mathcal{G} =(\mathcal{N},\mathcal{E}, \mathcal{W})$ composed of $\mathcal{N} = \{ N_1, ... , N_n\}$ nodes. They are connected by edges $\mathcal{E}= \{N_i-N_j : N_i \rightleftharpoons N_j \in \mathcal{E}\}$ with weights $w_{ij}=\mathcal{F}(i,j)\geq 0$ that represents a similarity measure $\mathcal{F}$ between the nodes $N_i\in \mathcal{N}$ and $N_j \in \mathcal{N}$. If $w_{ij}=0$ means that $(N_i, N_j)\notin \mathcal{E}$. In this work, a node represent an image in the graph.  This representation gives different advantages including strong mathematical properties, the ability to cope with annotation uncertainty and homogeneous space for highly heterogeneous data.

\smallskip
\textcolor{darkblue}{\textbf{Our All-in-One Framework.}} Semi-supervised classification has been explored from the model-based perspective in the medical domain e.g.~\cite{tiwari2010semi,zhao2014compact,wang2016progressive} or from the deep learning perspective using for example~\cite{gcn}.  Our framework lie in an different category: \textit{ hybrid techniques}, which seeks to keep the mathematical guarantees of model-based techniques whilst exploiting the power of deep nets. We remark that we use the term \textit{All-in-One} to describe the ability of our framework to plug-in different architectures and energy functionals without altering either the backbone nor the functionality  of our technique.

Our framework  has two  modes to operate: \textit{One Pass Classification} and \textit{Dynamic Pass Classification}. The key difference lies in the fact that the second option allows the uncertainty in the graph to be updated overtime. \textcolor{black}{The need for having two differing methods is as follows. For more complex datasets, iterative approaches are often needed to extract a rich feature representation. However, for simpler datasets, or those for which the training time is longer, a one-stop construction approach is the only computational feasible approach out of the two.} Both modes are composed of two main parts (i.e. hybrid model): i) a deep net, $f_\theta$ , that is used to generalise the feature extraction and reduce uncertainty in the labels and ii) a functional that seeks to diffuse the small amount of label data to the unlabelled set.

Our framework uses a given deep net, for example VGG16 or ResNet-18, defined as $f = \phi \circ \psi$, where $\psi : \mathcal{X}\rightarrow \mathbb{R}^P$ is a feature extractor and $\phi : \mathbb{R}^P \rightarrow \{0,1\}^c$ is the classification function. Firstly, our approach seek to solve one of the key problems in graph theory, which is -- how to construct an accurate embedding. 
\noindent\begin{minipage}{.48\textwidth}
{\begin{center}
\includegraphics[width=0.8\textwidth]{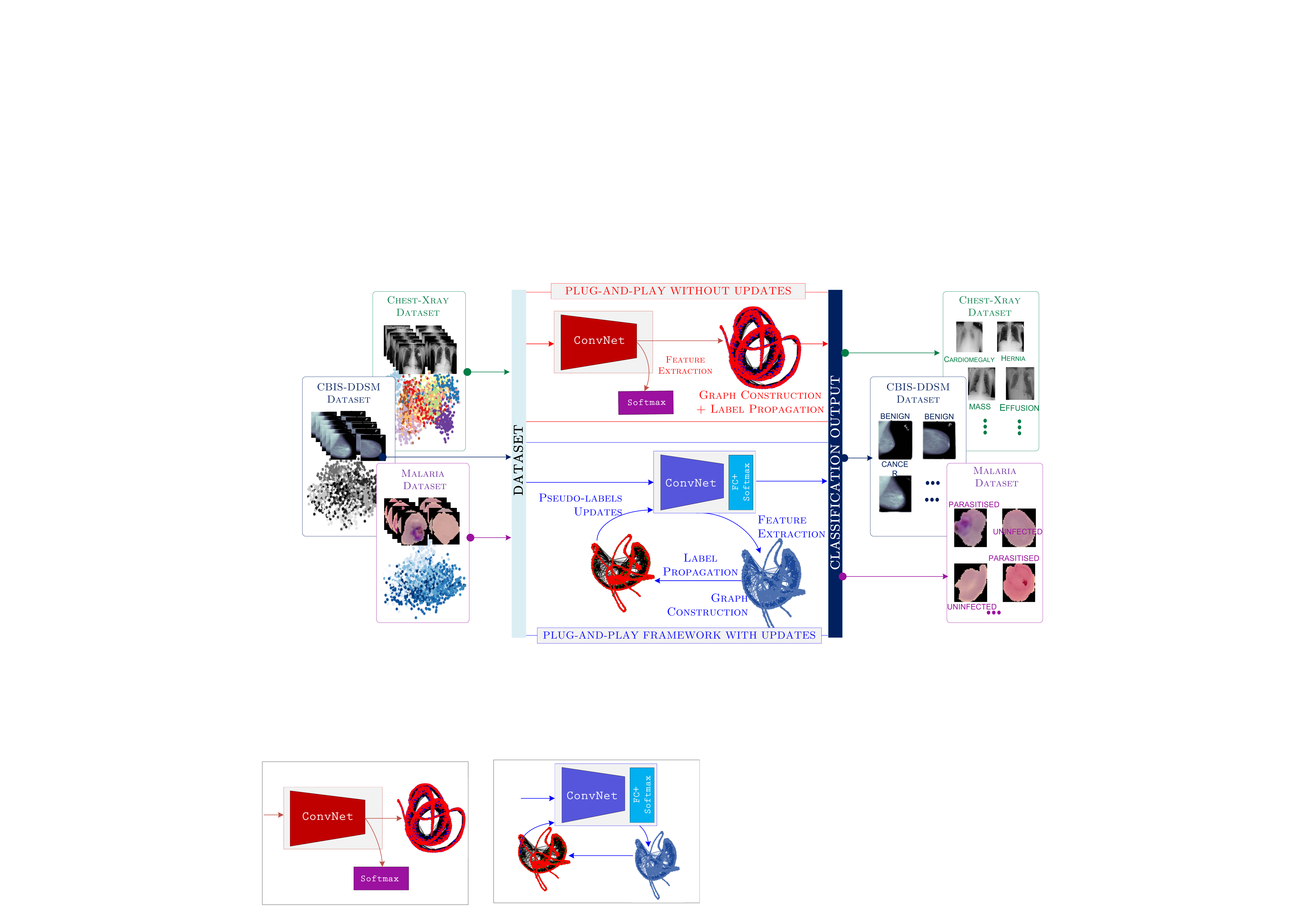}
\end{center}} \vspace{-0.7cm}
\begin{lstlisting}[caption=One Pass Classification,frame=tlrb, mathescape=true ]{Name}
Input: Data $X=\{x_1,..,x_n\}$, Labels $Y=\{y_1,..,y_l\}$ Architecture  $f_\theta(.) = \psi_\theta(.) \circ \phi_\theta(.)$

Optimise: $L_\theta = \sum_{i=i}^{l} l_s(f_{\theta}(x_i),y_i)$
#Generate features:
$N = \psi(X)$
#Construct weighted graph
$W_{ij} = d(n_i,n_j)$
$G=(V,E,W)$
#Compute label diffusion:
until convergence:
#Hypothesis class $\mathcal{H}$
$H^*=\text{argmin}_{H\in\mathcal{H}}\text{ }\mathbf{Q}(H) $  $$
#Extract generated labels:
$y_i = argmax_{k} h^{k}_i$
---
###energies included $\mathbf{Q}$:
## for the case of $p=2$
$S=D^{-1/2}WD^{-1/2}$
# given labeled set matrix Y
$H^{*}= (I-\alpha S)^{-1} Y$
## for the case of $p=1$
$\Delta_1(u)=|WD^{-1}u|$
minimise: $\sum_k \frac{\Delta_1 (u^k)}{|u^k|}$
\end{lstlisting}
\end{minipage}\hfill
\begin{minipage}{.48\textwidth}
{\begin{center}
\includegraphics[width=0.8\textwidth]{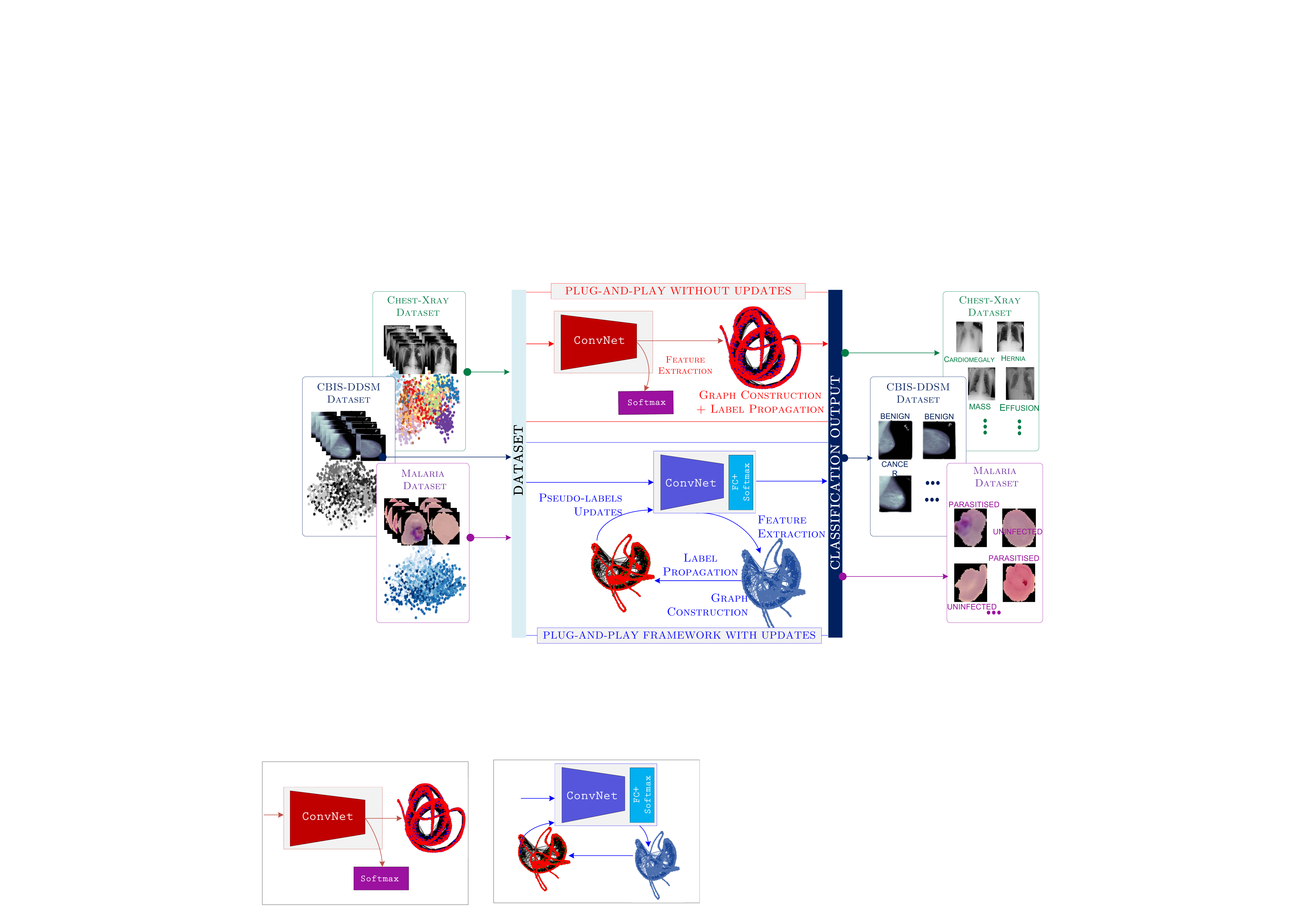}
\end{center}} \vspace{-0.5cm}
\begin{lstlisting}[caption=Dynamic Pass Classification,frame=tlrb,mathescape=true]{Name}
Input: Data $X=\{x_1,..,x_n\}$, Labels $Y=\{y_1,..,y_l\}$ Architecture  $f_\theta(.) = \psi_\theta(.) \circ \phi_\theta(.)$
Optimise: $L_\theta = \sum_{i=i}^{l} l_s(f_{\theta}(x_i),y_i)$
For J epochs:
 #Generate features:
 $N = \psi(X)$
 #Construct weighted graph
 $W_{ij} = d(n_i,n_j)$
 $G=(V,E,W)$
 #Compute label diffusion:
 until convergence:
 #Hypothesis class $\mathcal{H}$
 $H^*=argmin_{H\in\mathcal{H}}\text{ }\mathbf{Q}(H)$
 #Extract generated labels:
 $\hat{Y} = (\hat{y}_{l+1},..,\hat{y}_{n})$
 #$\alpha(t)$ balance parameter
 Optimise:
 $L_\theta = \sum_{i=i}^{l} l_s(f_{\theta}(x_i),y_i)$ $+ \alpha(t) \sum_{i=l+1}^{n} l_s(f_{\theta}(x_i),(\hat(y)_i)$
#Extract Final Generated Labels
$\hat{Y} = (\hat{y}_{l+1},..,\hat{y}_{n})$

\end{lstlisting}
\end{minipage}
For this, we extract the embeddings from a given deep net to better generalise the feature space to construct a graph. For the second part (i.e. the label diffusion), our setting is focused on the normalised graph p-Laplacian $\Delta_p(u)$. Whilst $p=2$ has been used extensively in the literature e.g.~\cite{chen2013inferring,dodero2014group} other more robust functionals have not been deeply explored in the medical domain. More recently, authors of that~\cite{aviles2019graphx} explored a more robust functional based on the case for $p=1$ $\Delta_1(u) =  |WD^{-1}u|$, where $W$ is the weight matrix and $D$ a matrix with the node's degrees. We show in this work, that we can plug-in any functional in our framework. Our GraphNet Zoo includes energies based on $p=2$ e.g.~\cite{zhou2004learning} and more robust ones based on the $p=1$ case e.g.~\cite{aviles2019graphx}. We explicitly define the process of our framework below and set explicitly the energies used in this work (see lines 16-23 from the algorithm).

\textcolor{darkblue}{\textbf{\textit{One-Pass Graph Classification}}}: This mode allows us to perform deep semi-supervised classification based on the conditional entropy of the class probabilities for the unlabelled set. To train a deep net in a SSL fashion, we rely on \textit{pseudo-labels}~\cite{grandvalet2005semi,lee2013pseudo} where the key idea is to approximate the class label for the unlabelled set from the predictions of a deep net $f_\theta$. This allows to strengthen the graph constructions whilst boosting the classification performance. In short, we extract the embeddings from a deep net trained over the labelled set to build the graph representation of the data and finally apply label diffusion $\mathbf{Q}(H)$. The precise optimisation process is displayed in lines 1:23 on Listing 1.1.

\textcolor{darkblue}{\textbf{\textit{Dynamic Pass Graph Classification}}}: This mode seeks to improve the uncertainty generated by the pseudo-labels overtime. We observe that whilst the first alternative offers good results in terms of classification, there is two ways to further improve it. Firstly the inferred pseudo-labels  clearly do not have the same confidence in each example, and secondly the pseudo-labels may be imbalanced over the classes which affect the learning process. We tackle these problems by associating, to each pseudo-label, a weight reflection the inference certainty. We use, as in other works~\cite{iscen2019label,sellars2020two}, an entropy measure $M : \mathbb{R}^{C} \rightarrow \mathbb{R}$ to assign the certainty $\xi_i= 1 - (M(h^*_i)/log(C))$ to a given example $x_i$. Note that $h^*_i$ comes from line: 25 of the below algorithm whilst $C$ denotes the classes. The overall procedure is described in Listing 1.2 (lines 17:42).

We detail the implementation of the two approaches in the Listing 1.1 and Listing 1.2, which are displayed next.

\section{Experimental Results}
This section is devoted to give details of the experiments carried out to validate our proposed framework.

\smallskip
\noindent
\textcolor{darkblue}{\textbf{Data Description.}} We evaluate our approach on three large scale datasets. The first one is \textit{the Malaria Infected Cells}~\cite{malaria}  dataset. It is composed of 27,558 cell images of size $224 \times 224$ with balance instances of parasitised  and uninfected cells.
As second dataset, we use the latest version of the very challenging  \textit{Digital Database for Screening Mammography (DDSM)} dataset called CBIS-DDSM~\cite{lee2017curated}. It has 2478 mammography images from 1249 women the images  size can be as large as 3k~$\times$~3k pixels. It is composed of roughly  40:60 of benign and malignant cases respectively. Finally, we use  the ChestX-ray14 dataset \cite{wang2017chestx}, which is composed of 112,120 frontal chest view X-ray with size of 1024$\times$1024. The dataset is composed of 14 classes (pathologies).

\begin{table}[t!]
    \begin{minipage}{.5\linewidth}
      \centering
       \resizebox{1\textwidth}{!}{
        \begin{tabular}{cclllll}
            \hline
            \multicolumn{7}{c}{\cellcolor[HTML]{EFEFEF} \textsc{Malaria Cells Dataset}} \\ \hline
            \multicolumn{7}{c}{\textsc{Fully-Supervised Methods}} \\ \hline
            \multicolumn{1}{c}{\textsc{Method}} & \multicolumn{6}{c}{\textsc{Accuracy} (Labelled 70\%) } \\ \hline
            Xception & \multicolumn{6}{c}{0.890} \\
            VGG-16 & \multicolumn{6}{c}{\cellcolor[HTML]{F1F8E9}0.945} \\
            ResNet-50 & \multicolumn{6}{c}{\cellcolor[HTML]{DCEDC8}0.957} \\
            AlexNet & \multicolumn{6}{c}{0.937} \\ \hline\hline
            \multicolumn{7}{c}{\begin{tabular}[c]{@{}c@{}} \textsc{Deep Semi-Supervised Models}\end{tabular}} \\ \hline
            \multicolumn{1}{c}{\textsc{Method}} & 10\% & \multicolumn{5}{c}{20\%} \\ \hline
            \multicolumn{1}{c}{GCN~\cite{kipf2016semi}} & \multicolumn{1}{l}{ 0.865$\pm$ 0.05 } & \multicolumn{5}{l}{0.895$\pm$ 0.015} \\
            \multicolumn{1}{c}{\textcolor{blue}{Ours W/[A]}} & \multicolumn{1}{c}{0.877 $\pm$0.006} & \multicolumn{5}{c}{0.927 $\pm$ 0.005} \\
            \multicolumn{1}{c}{\textcolor{blue}{Ours W/[B]} } & \multicolumn{1}{c}{0.930$\pm$ 0.009} & \multicolumn{5}{c}{\cellcolor[HTML]{F1F8E9}0.943 $\pm$0.005} \\
            \multicolumn{1}{c}{\textcolor{blue}{Ours W/[C]}} & \multicolumn{1}{c}{0.845 $\pm$0.034} & \multicolumn{5}{c}{0.921$\pm$ 0.011} \\
            \multicolumn{1}{c}{\textcolor{red}{Ours W/[A]}} & \multicolumn{1}{c}{0.922 $\pm$0.004} & \multicolumn{5}{c}{0.928$\pm$ 0.009} \\
            \multicolumn{1}{c}{\textcolor{red}{Ours W/[B]}} & \multicolumn{1}{c}{\cellcolor[HTML]{F1F8E9}0.94 $\pm$0.0057} & \multicolumn{5}{c}{\cellcolor[HTML]{DCEDC8}	0.957$\pm$ 0.003} \\
            \multicolumn{1}{c}{\textcolor{red}{Ours W/[C]}} & \multicolumn{1}{c}{0.860 $\pm$0.038} & \multicolumn{5}{c}{0.929$\pm$ 0.010} \\ \hline
      \end{tabular}
      }
    \end{minipage}%
    \begin{minipage}{.5\linewidth}
      \centering
      { \scriptsize{ \textsc{Colour Code and Architectures:}

      \smallskip
      \textcolor{blue}{GraphNet Zoo}: Model 1;
      \textcolor{red}{GraphNet Zoo}: Model 2}
      \smallskip

      [A]: VGG16; [B]:ResNet-18; [C]: AE

      [D]: Custumised CNN; [E]: ResNet-50
      }
      \\
      \bigskip
        \resizebox{1\textwidth}{!}{
        \begin{tabular}{cclllll}
            \hline
            \multicolumn{7}{c}{\cellcolor[HTML]{EFEFEF} \textsc{CBIS-DDSM dataset}} \\ \hline
            \multicolumn{7}{c}{\textsc{Fully-Supervised Methods}} \\ \hline
            \multicolumn{1}{c}{\textsc{Method}} & \multicolumn{6}{c}{\textsc{[\% Labelled] AUC } } \\ \hline
            \textsc{Shen}~\cite{shen2019deep} & \multicolumn{6}{c}{[85\%] \cellcolor[HTML]{DCEDC8} 0.85 (val) \textbf{0.75 (test)}} \\
            \textsc{Zhu}~\cite{zhu2017deep} & \multicolumn{6}{c}{ [80\%] 0.791} \\
             \hline\hline
            \multicolumn{1}{c}{\textsc{Method}} & \multicolumn{6}{c}{\textsc{20\% Labelled -- AUC} } \\ \hline
            \multicolumn{1}{c}{\textcolor{blue}{Ours W/[D]}} & \multicolumn{6}{c}{\cellcolor[HTML]{F1F8E9}0.729} \\
            \multicolumn{1}{c}{\textcolor{blue}{Ours W/[E]}} & \multicolumn{6}{c}{0.717} \\
            \multicolumn{1}{c}{\textcolor{red}{Ours W/[D]}} & \multicolumn{6}{c}{0.721} \\
            \multicolumn{1}{c}{\textcolor{red}{Ours W/[E]}} & \multicolumn{6}{c}{\cellcolor[HTML]{DCEDC8}0.735 } \\  \hline
            \multicolumn{1}{c}{\textcolor{red}{Ours W/[E]}} & \multicolumn{6}{c}{\cellcolor[HTML]{FFFFC7} [40\%] 0.811 } \\
            \hline\hline
     \end{tabular}
     }
    \end{minipage}
    \vspace{0.2cm}
\caption{Performance comparison on the Malaria  and Mammogram datasets. For both datasets GraphNet Zoo was able to produce state-of-the-art performance, beating the compared methods, whilst using far fewer labelled data points. }
\label{table1}
\vspace{-0.6cm}
\end{table}

\smallskip
\noindent
\textcolor{darkblue}{\textbf{Parameter Selection.}} \textit{Architectures:} We set the learning rate to the common value of $r=0.001$ in all our networks. Additionally, we use the Adam optimiser with early stopping and a dropout rate of 0.2. \textit{Graph:} For the graph based approaches we used k-NN for the graph construction with k=50, $\alpha$ was set with a grid search in $[0,1]$. Moreover, we used $p=2$ based energy for the malaria and the mammograms dataset whilst for the chest-xray we used the label diffusion based on minimising our energy based $p=1$. \textbf{Data Pre-processing:} We follow standard pre-processing protocol (e.g. as in~\cite{LiShen2017}) to normalise the images so that the mean of the pixel values is 1 and the standard deviation is 1. For the compared approaches, we used the code provided by each author along with their parameters.  

\begin{figure*}[!t]
\centering
\includegraphics[width=1\textwidth]{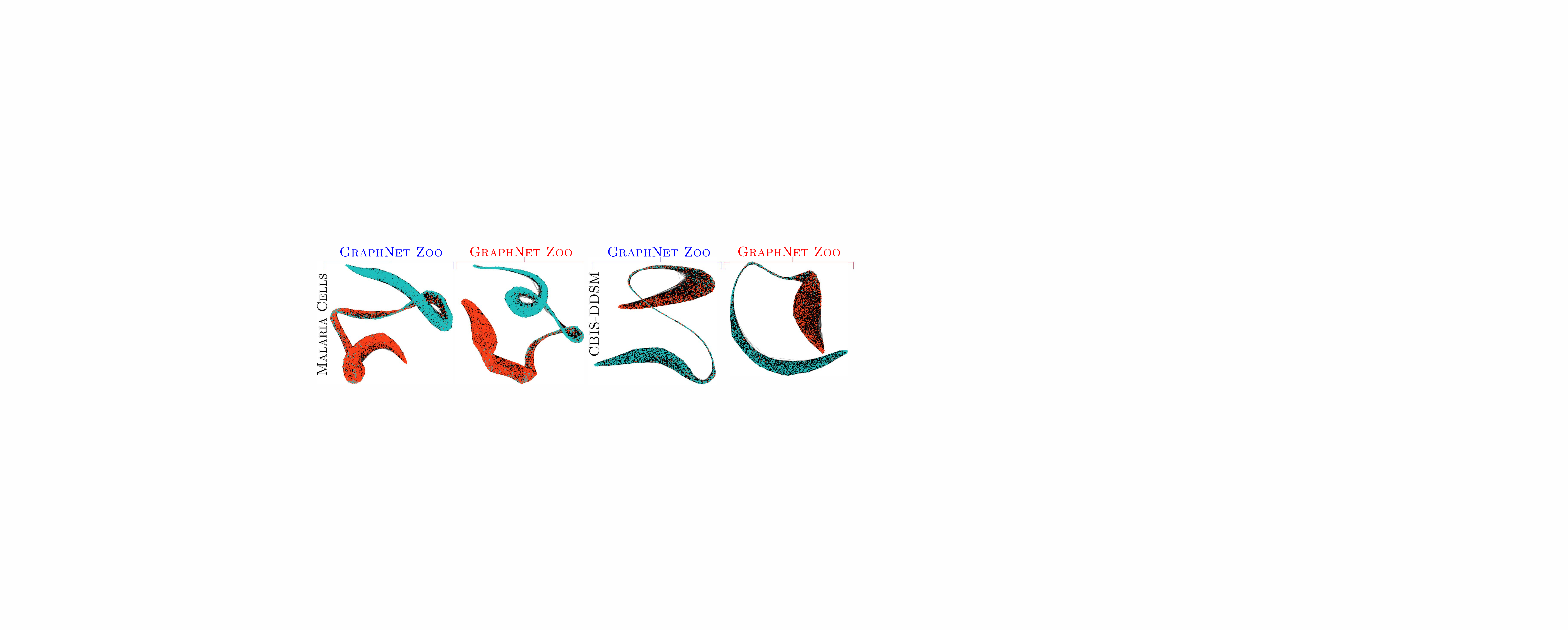}
\caption{Feature representation for the Malaria \cite{malaria} and Mammogram \cite{lee2017curated} datasets using our graph based approaches using a ResNet18 feature extractor. Red dots denote samples labeled malignant and blue denotes benign samples.  }
\label{fig::visu1}
\end{figure*}


\begin{figure*}[!t]
\centering
\includegraphics[width=1\textwidth]{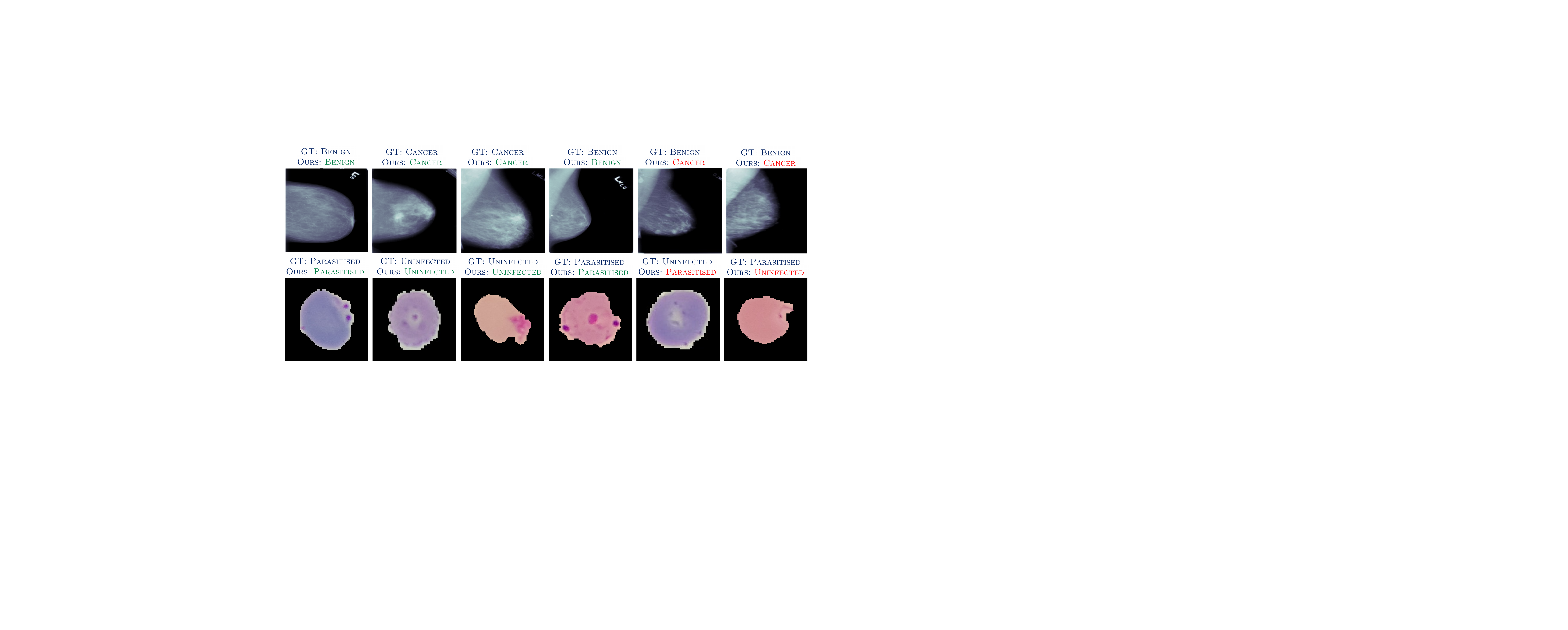}
\caption{Example classifications from the Malaria \cite{malaria} and Mammogram \cite{lee2017curated} datasets. For each image we give the ground truth class and the predicted classification. Whilst our approach works very well, there are still miss-classifications, as in any algorithmic approach for such applications.}
\label{fig::visu1}
\end{figure*}

\smallskip
\noindent
\textcolor{darkblue}{\textbf{Evaluation Protocol.}} We use the following evaluation scheme to validate our framework. We validate our approach by comparing our deep semi-supervised framework against fully-supervised SOTA-techniques for each particular application including the works of that ~\cite{zhu2017deep,shen2019deep,wang2017chestx,yao2018weakly,yan2018weakly,baltruschat2019comparison}. Moreover, we also compared our framework against the  semi-supervised techniques of ~\cite{gcn,aviles2019graphx}. As performance check we use two metrics: accuracy and a receiver operating characteristic curve (ROC) analysis based on the Area Under the Curve (AUC).

\smallskip
\noindent
\textcolor{darkblue}{\textbf{Results and Discussion.}} We begin by analysing the Malaria \cite{malaria} and Mammography \cite{lee2017curated} datasets, see Table \ref{table1}. As our PnP framework is versatile, we can apply any combination of feature extractor and graphical classifier, which can be swapped in and out. To show this ability, we ran our framework using a range of different architectures and graphical propagators. From a closer look at the results, we notice that compared to the other SSL approaches the GCN~\cite{kipf2016semi} produces consistently lower results, which suggests that the shallow architecture of GCN's generalise poorly to more complex datasets. However, our framework gives far better results than GCN. We further support our results by running a comparison against fully supervised methods for each particular application. For each dataset we learnt on a very small amount of labelled data but yet managed to obtain a great classification performance which was either comparable to or better than the state-of-the-art supervised approaches which used far more labels. The fact that our approach worked so well for the CBIS-DDSM dataset is of particular interest as it presents a challenging dataset but yet we were
able to get similar performance to SOTA techniques using $20\%$ of the label set but using $40\%$ we were able to surpass all comparison methods using half the label set. This could be explained by noisy labels, uncertainty from the experts, contributing negatively whilst training in a  supervised manner but can be regularised against using semi-supervised approaches.

To test \textit{the generalisability} of our framework we then applied it to the challenging ChestXray-14 dataset~\cite{wang2017chestx}, and the results are reported in Table 2 and Fig 4. From the results, one can observe that using far less labels we are able to match or outperform the SOTA-methods.
Overall, we underline the message of this paper \textit{we show that our framework is easy applied to a variety of problems and that it reliable produces good performance using fewer labels.  }

\begin{table}[t!]
    \begin{minipage}{.6\linewidth}
      \centering
       \resizebox{1\textwidth}{!}{
\begin{tabular}{cclllll}
\hline
\multicolumn{7}{c}{\cellcolor[HTML]{EFEFEF} \textsc{ChestXray-14}} \\ \hline
\multicolumn{7}{c}{\textsc{Fully-Supervised Methods}} \\ \hline
\multicolumn{1}{c}{\textsc{Method}} & \multicolumn{6}{c}{\textsc{70\% Labelled -- AUC} } \\ \hline
\textsc{Wang~\cite{wang2017chestx}} & \multicolumn{6}{c}{0.745} \\
\textsc{Yao~\cite{yao2018weakly}} & \multicolumn{6}{c}{0.761} \\
\textsc{Yan~\cite{yan2018weakly}} & \multicolumn{6}{c}{\cellcolor[HTML]{DCEDC8}0.830} \\
\textsc{Baltruschat (ResNet-101)~\cite{baltruschat2019comparison}} & \multicolumn{6}{c}{\cellcolor[HTML]{F1F8E9}0.785} \\
\textsc{Baltruschat (ResNet-38)~\cite{baltruschat2019comparison}} & \multicolumn{6}{c}{\cellcolor[HTML]{DCEDC8} 0.806} \\
 \hline\hline
\multicolumn{1}{c}{\textsc{Method}} & \multicolumn{6}{c}{\textsc{20\% Labelled -- AUC} } \\ \hline
\multicolumn{1}{c}{GraphXNet~\cite{aviles2019graphx}} & \multicolumn{6}{c}{0.788} \\
\multicolumn{1}{c}{\textcolor{blue}{Ours W/[A1]}} & \multicolumn{6}{c}{0.770} \\
\multicolumn{1}{c}{\textcolor{red}{Ours W/[A1]}} & \multicolumn{6}{c}{\cellcolor[HTML]{F1F8E9} 0.795} \\
\multicolumn{1}{c}{\textcolor{red}{Ours W/[A2]}} & \multicolumn{6}{c}{\cellcolor[HTML]{DCEDC8} 0.815} \\
\hline\hline
\end{tabular}}
\label{table2}
\caption{Classification performance on the ChestXray-14 dataset. When only $20\%$ of the dataset labelled we are able to beat  and perform in line with recent methods. }
\end{minipage}
\begin{minipage}{.35\linewidth}
\def\tablename{Fig. 4. }
\centering
\includegraphics[width=1\textwidth]{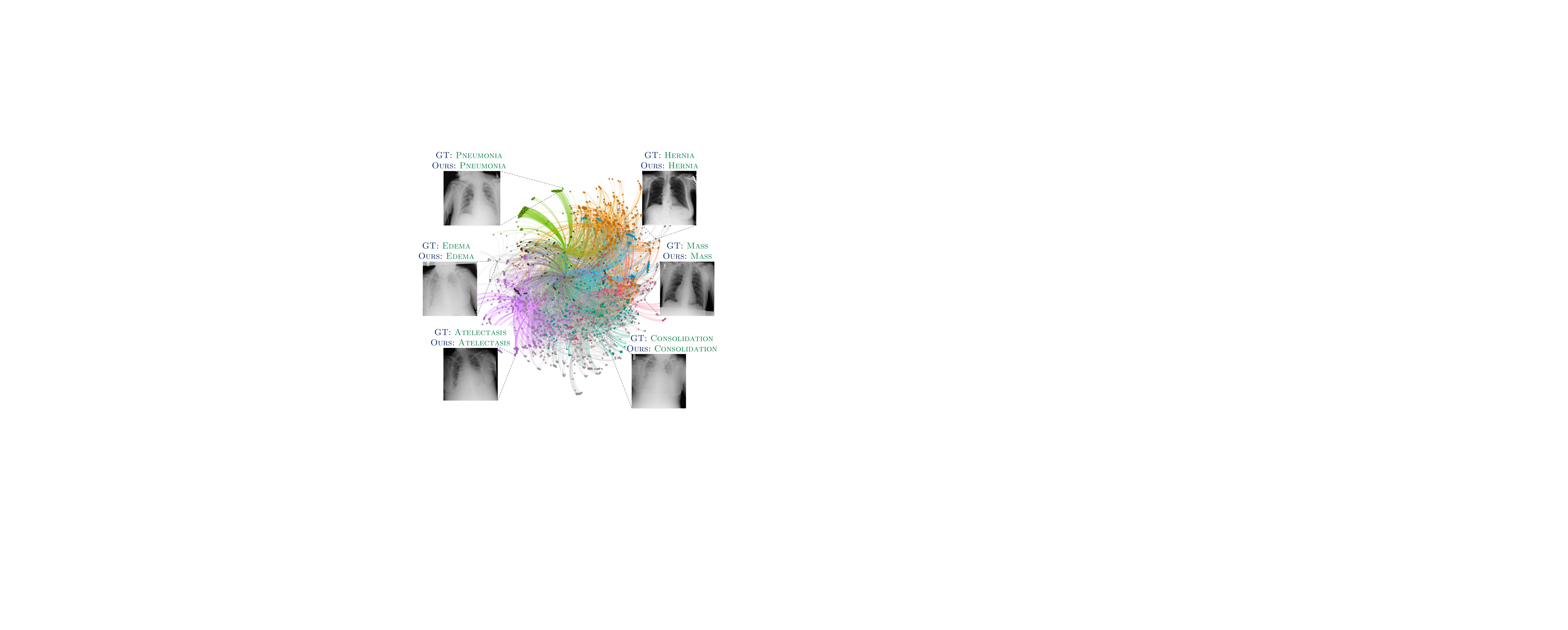}
\caption*{Fig. 4: Visual display of one of our experiments: a graph output in which each colour represent one pathology (class).}
\label{fig::visu1}
\end{minipage}
\vspace{-0.6cm}
\end{table}


\section{Conclusion}
In the field of medical imaging labelled examples are time consuming and expensive to obtain. Supervised approaches often rely upon a large representative training set to achieve acceptable performance. In this paper, we explore the impact that semi-supervised learning (SSL) can have in the domain. We propose a novel framework for SSL algorithms before applying this framework to three large scale medical datasets. Through extensive testing, we clearly show that our graph-based approach can either match or outperform state-of-the-art deep supervised methods whilst requiring a fraction of the labels -- only 20\%.
Overall, \textit{we underline the message of our paper, deep SSL classification is reaching unprecedented performance comparable or better than fully-supervised techniques whilst requiring minimal labelled set.}
%
%
\bibliographystyle{splncs04}
\bibliography{bibMICCAI20.bib}

\end{document}